\documentclass[letterpaper, 10 pt, conference]{ieeeconf}  

\IEEEoverridecommandlockouts                              

\usepackage{graphicx}
\usepackage[hidelinks]{hyperref}

\usepackage{color,soul}
\usepackage[usenames,dvipsnames]{xcolor}
\usepackage{caption}
\usepackage{subcaption}
\usepackage{multirow}
\usepackage{array}
\usepackage{booktabs}
\usepackage{graphicx}
\usepackage{caption}
\usepackage{tabularx}
\usepackage{float}
\usepackage[normalem]{ulem}
\usepackage{makecell}
\usepackage{placeins}
\usepackage{fancyhdr}

\newcommand{\ieeecopyright}{%
  \footnotesize
  \centering
}

\fancypagestyle{ieeefirstpage}{%
  \fancyhf{}
  \fancyfoot[C]{\ieeecopyright}
}

\title{\LARGE \bf
\SYS{}: Parallel Loop Closing with GPU-Acceleration in Visual SLAM
\vspace{-.5cm}
}

\author{Soudabeh Mohammadhashemi$^{1}$, Shishir Gopinath$^{1}$, Kimia Khabiri$^{1}$, 
\\ Parsa Hosseininejad$^{1}$, Karthik Dantu$^{2}$, Steven Y. Ko$^{1}$
\thanks{$^{1}$Simon Fraser University. 
\textcolor{black}\texttt{\{
\href{mailto:sma406@sfu.ca}{\textcolor{black}{sma406}},
\href{mailto:sgopinat@sfu.ca}{\textcolor{black}{sgopinat}},
\href{mailto:kka156@sfu.ca}{\textcolor{black}{kka156}},
\href{mailto:sph6@sfu.ca}{\textcolor{black}{sph6}}, \href{mailto:steveyko@sfu.ca}{\textcolor{black}{steveyko}}\}@sfu.ca}}
\thanks{$^{2}$University at Buffalo. 
\texttt{\href{mailto:kdantu@buffalo.edu}{\textcolor{black}{kdantu@buffalo.edu}}}}
}

\newcommand{\SYS}[1]{\textcolor{black} {FastLoop}}
\newcommand{\TUMVI}[1]{\textcolor{black} {TUM-VI}}
\newcommand{\EUROC}[1]{\textcolor{black} {EuRoC}}

\begin{document}

\bstctlcite{IEEEexample:BSTcontrol}

\maketitle
\thispagestyle{ieeefirstpage}
\pagestyle{empty}

\begin{abstract}
Visual SLAM systems combine visual tracking with global loop closure to maintain a consistent map and accurate localization. Loop closure is a computationally expensive process, as we need to search across the whole map for matches. This paper presents \SYS{}, a GPU-accelerated loop closing module to alleviate this computational complexity. We identify key performance bottlenecks in the loop closing pipeline of visual SLAM and address them through parallel optimizations on the GPU. Specifically, we use task-level and data-level parallelism and integrate a GPU-accelerated pose graph optimization. Our implementation is built on top of ORB-SLAM3 and leverages CUDA for GPU programming. Experimental results show that, on the embedded platform, \SYS{} achieves an average speedup of $1.3\times$ on the \EUROC{} dataset and $2.4\times$ on the \TUMVI{} dataset. On the desktop platform, it achieves average speedups of $1.4\times$ and $3.0\times$ on the \EUROC{} and \TUMVI{} datasets, respectively, while maintaining the accuracy of the original system.

\end{abstract}

\section{Introduction} \label{sec:introduction}

Simultaneous Localization and Mapping (SLAM) enables robots to estimate their pose while incrementally building a representation of their surrounding environment, using data from multiple sensors such as cameras, LiDAR, and inertial measurement units (IMUs). Among these, Visual SLAM relies specifically on visual information and typically employs monocular, stereo, or RGB-D cameras to reconstruct the environment and localize the robot within it.

A typical visual SLAM system consists of three main components: tracking, local mapping, and loop closing~\cite{Semenova2024}. The tracking module estimates the camera pose by detecting features in incoming frames and matching them to previously mapped landmarks. Local mapping incrementally updates and optimizes the map to improve overall accuracy, while loop closing identifies revisited locations to correct accumulated drift. Well-known systems following this architecture include ORB-SLAM~\cite{Mur-Artal2015, Mur-Artal2017, Campos2021} and Kimera~\cite{Kimera, kimera2021}.

In practice, a SLAM system first performs visual odometry and then applies loop closure to mitigate drift when revisiting previously seen locations. Cameras typically capture images at 30~fps, and most LiDARs provide scans at around 10~Hz ~\cite{Kumar2024}. To maintain real-time operation, it is essential that all processing is performed promptly, including loop closure.
Though it occurs rarely as it is triggered only when a loop is detected, when it does, 
its execution time can increase substantially compared to a regular iteration. For instance, place recognition alone typically takes on the order of 10\,ms, while the subsequent map correction and full map optimization steps can take on the order of hundreds of milliseconds to over a second, depending on map size \cite{Campos2021, picard2023surveyrealtime3dscene}. Such latency spikes have been shown to stall the tracking and mapping threads long enough to cause temporary tracking loss on resource-constrained platforms \cite{Aldegheri2019}. This makes loop closing a disruptive bottleneck despite its infrequency, particularly on hardware with limited resources.

Reducing this latency while sustaining real-time performance is therefore essential, and one promising solution is parallelization.
Studies have leveraged GPU acceleration to speed up various SLAM components, including tracking~\cite{Khabiri2025, AdrianR, KumarParkBehera2024} and mapping-related tasks such as bundle adjustment~\cite{korovko2025cuvslamcudaacceleratedvisual}. However, none of these efforts target loop closing, despite it being executed for every newly inserted keyframe in the map and, as discussed above, becoming a major source of latency spikes when a loop is detected.

However, applying GPU acceleration to loop closing
is challenging for two main reasons. First, the three main components of modern visual SLAM systems, which are tracking, local mapping, and loop closing, concurrently access and modify shared data structures, including the map features like keyframes, map points, and graphs. For example, loop closing operations may update keyframe poses, fuse map points, or change graph connections. Running these operations in parallel without careful coordination can cause conflicts and produce incorrect results. Second, several computational patterns 
are repeatedly executed, including repeated feature association and map point fusion operations.
When executing on the GPU, each execution requires transferring the relevant data structures and launching dedicated GPU tasks (i.e., kernels) to process them. This leads to frequent CPU–GPU data transfers and redundant kernel launches, ultimately diminishing the potential performance gains.

To overcome these challenges, we propose a new loop closing technique named \SYS{} in visual SLAM systems. We analyze the loop closing pipeline's workload and data flow and observe that several components inherently exhibit task-level or data-level parallelism that make them amenable to parallel execution. While task- and data-level parallelism are well-established techniques, the difficulty is not in the concepts themselves, but in recognizing where they apply within the loop closing pipeline and in redesigning its internals so that exploiting them yields an actual performance gain rather than being offset by transfer and synchronization overhead.
Thus, our approach (1) identifies the specific opportunities for parallelization present in the loop closure computation, (2) classifies each opportunity according to whether it exhibits task-level or data-level parallelism, and (3) redesigns the pipeline's internal computation and data flow to expose and exploit these opportunities and reduce data transfer overhead.
By combining these strategies, we transform a traditionally sequential loop closing pipeline into a parallel architecture.


Building on these ideas, we design and implement \SYS{}, which makes the following contributions: (1) \textbf{Parallelizing independent per-unit computations.} We observe that certain loop closure components, such as loop fusion and projection search, repeat the same computation independently across many map points or keyframes, with no dependencies between units. We exploit this data-level parallelism by assigning one GPU thread to each independent unit. (2) \textbf{Parallelizing dependency-free components.} We observe that other components consume the same input data but do not depend on one another's output. We exploit this task-level parallelism by executing them concurrently on the GPU instead of sequentially as in the original pipeline.
(3) \textbf{Reduced CPU--GPU transfer overhead.} We reduce data transfer overhead by transferring data to the GPU only once and reusing it across task-parallel instances, combined with pinned memory for faster host-to-device transfers.
Together, these contributions enable substantial reductions in loop closing latency without sacrificing trajectory accuracy.

We integrate \SYS{} into ORB-SLAM3 as a representative visual SLAM system to demonstrate the benefits of our approach. Performance evaluation is conducted using \EUROC{}~\cite{euroc} and \TUMVI{}~\cite{tumvi} datasets. Our experiments show that \SYS{} achieves an average speedup of $1.5\times$ and $1.4\times$ on \EUROC{} and $3.1\times$ and $2.4\times$ on \TUMVI{} for the loop closing module on both desktop and embedded platforms, respectively, while maintaining comparable trajectory accuracy. Although our implementation is based on ORB-SLAM3, the proposed techniques are expected to generalize to other visual SLAM systems that maintain a global map by performing global loop closure.
Our code is available at \url{https://github.com/sfu-rsl/FastLoop}.

\section{Related Work} \label{sec:related-work}

In this section, we briefly review related work on GPU acceleration for various components of visual SLAM systems.

Many approaches accelerate frontend tracking components of visual SLAM systems.
Aldegheri et al.~\cite{Aldegheri2019} accelerate feature extraction by exploiting data flow in ORB-SLAM2.
Muzzini et al.~\cite{Muzzini2024} design high-performance GPU-accelerated ORB feature extraction techniques and achieve up to a $3\times$ speedup compared to previous methods. Kumar et al.~\cite{KumarParkBehera2024} propose a new front-end-middle-end design, which improves visual tracking performance by using bounded rectification, along with parallelized pyramidal culling and aggregation. 
Khabiri et al.~\cite{Khabiri2025} identify and accelerate additional time-consuming tracking tasks such as stereo matching and local map tracking in ORB-SLAM3.

Meanwhile, others have focused on accelerating mapping-related tasks. Prior works accelerate bundle adjustment~\cite{cudabundleadjustment, Gopinath2023} in ORB-SLAM-based systems. Lu et al.~\cite{Lu2022} propose a modified VINS-Mono algorithm which accelerates nonlinear least-squares and marginalization using CUDA. 
Hosseininejad et al.~\cite{hosseininejad2026turbomapgpuacceleratedlocalmapping} offload local mapping tasks to the GPU by restructuring the logic of map point creation and fusion, while also accelerating redundant keyframe culling on the CPU in ORB-SLAM3. Korovko et al.~\cite{korovko2025cuvslamcudaacceleratedvisual} also propose a CUDA-accelerated SLAM system to improve the performance of bundle adjustment, feature tracking, and several other components. 

In contrast, the acceleration of loop closure using a GPU is largely unexplored, despite it being critical for correcting accumulated errors in SLAM. Modern SLAM systems~\cite{Campos2021, korovko2025cuvslamcudaacceleratedvisual} decouple tracking from loop closing so that expensive operations do not slow down time-sensitive input processing. However, these systems do not apply GPU acceleration to pose graph optimization, a higher-latency step in loop correction.
An et al.~\cite{An2019} develop a GPU-based loop detection method using convolutional neural networks. Kumar et al.~\cite{Kumar2024} propose a GPU-accelerated numerical differentiation method to speed up pose graph optimization inside of ORB-SLAM3. However, in this approach, several steps are still executed on the CPU, which involves multiple GPU-CPU data transfers. In our work, we develop a new pose graph implementation using a GPU-optimized library to mitigate this.




Building on prior work, we focus on parallelizing the most time-consuming components of the loop closing module on the GPU. In particular, we leverage both task-level and data-level parallelism and employ GPU-accelerated graph optimization using automatic differentiation to improve the performance of loop closing.

\section{Background} \label{sec:background}

\begin{figure}[t]
    \centering
    \includegraphics[width=0.8\linewidth]{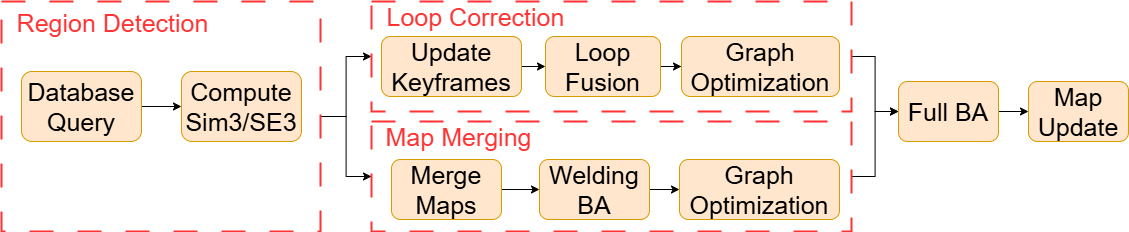}
    \caption{Loop Closing workflow in ORB-SLAM3.}
    \label{fig:loop_closing_stages}
    \vspace{-15pt}
\end{figure}

Many Visual SLAM systems includes a module called loop closure, which is responsible for 
detecting if the current keyframe is similar to any other location previously visited. 
When a common region is found within the map, loop correction is performed. 
As part of the loop correction, a full bundle adjustment (BA) is executed to refine the map.

\autoref{fig:loop_closing_stages} illustrates the various stages of this module, taking ORB-SLAM3 as an example. As described earlier, region detection can be performed very fast. For this reason, we focus solely on loop correction. 

\subsection{Region Detection}
\label{bg:place_recognition}

In ORB-SLAM3, region detection is triggered whenever the mapping component creates a new keyframe and aims to find matches with keyframes stored in the map. If the matched keyframe belongs to the active map, a loop correction is performed. Keyframe matching is performed using visual bag of words matching, which helps to efficiently retrieve the most similar candidates for a given query image. After estimating the relative pose between the new keyframe and the matched one, a local window is defined using the matched keyframe and its neighbors in the covisibility graph to ensure consistent verification.



\subsection{Loop Correction}
\label{bg:loop_correction}
When a successful region detection establishes a data association between the current keyframe and a previously visited keyframe in the same map, the loop correction procedure is triggered. The process starts by updating the connections between the current keyframe and its connected keyframes, followed by correcting their poses using the estimated Sim3 transformation and updated velocity. To ensure global consistency, a loop fusion step is then performed, where a window is constructed around the matched keyframes to identify and merge duplicate map points based on geometric and descriptor similarity. This fusion removes redundant landmarks and creates new connections in the covisibility and essential graphs, resulting in a cleaner and more consistent map representation. Finally, pose graph optimization over the essential graph propagates the loop correction to the rest of the map, enforcing global consistency across all keyframes. 



\begin{figure}[t]
    \centering
    \includegraphics[width=0.6\linewidth]{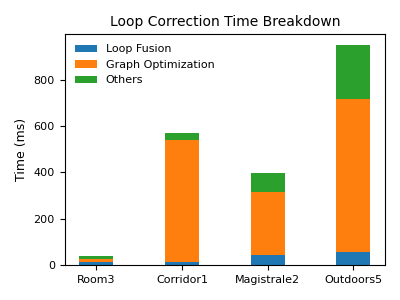}
    \caption{Average time spent in different sections of Loop Correction in ORB-SLAM3.
    }
    \label{fig:loop_correction_time_breakdown}
    \vspace{-15pt}
\end{figure}

\begin{figure*}[!t]
    \centering
    \begin{subfigure}[t]{0.8\linewidth}
        \centering
        \includegraphics[width=\linewidth]{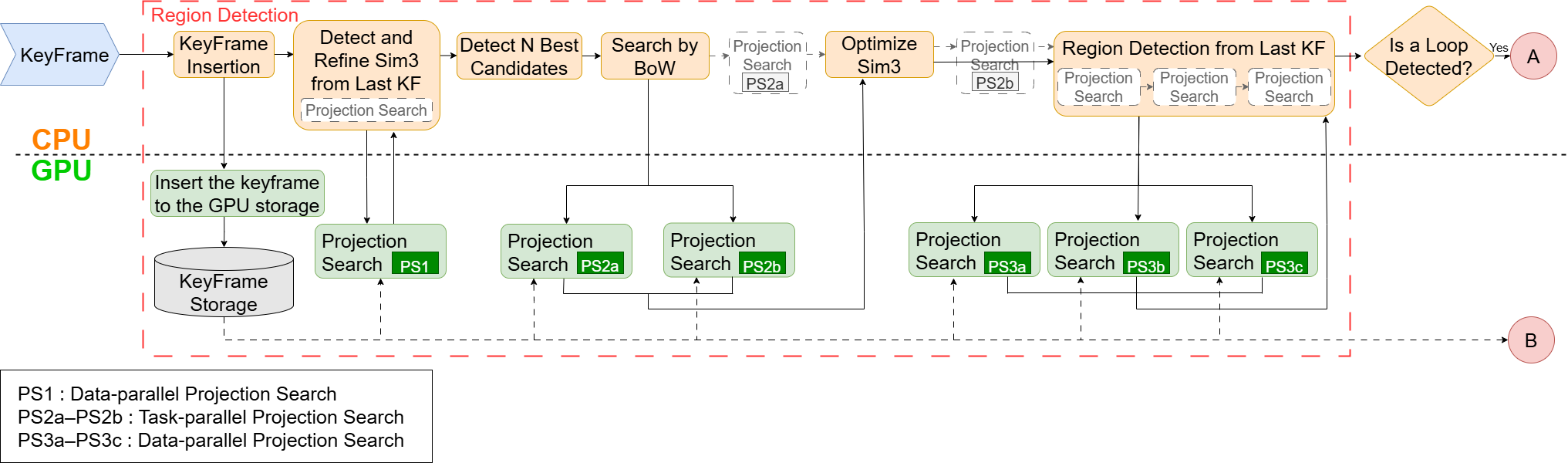}
        \caption{Region Detection components}
        \label{fig:fastLoop_breakdown_a}
    \end{subfigure}


    \begin{subfigure}[t]{0.8\linewidth}
        \centering
        \includegraphics[width=\linewidth]{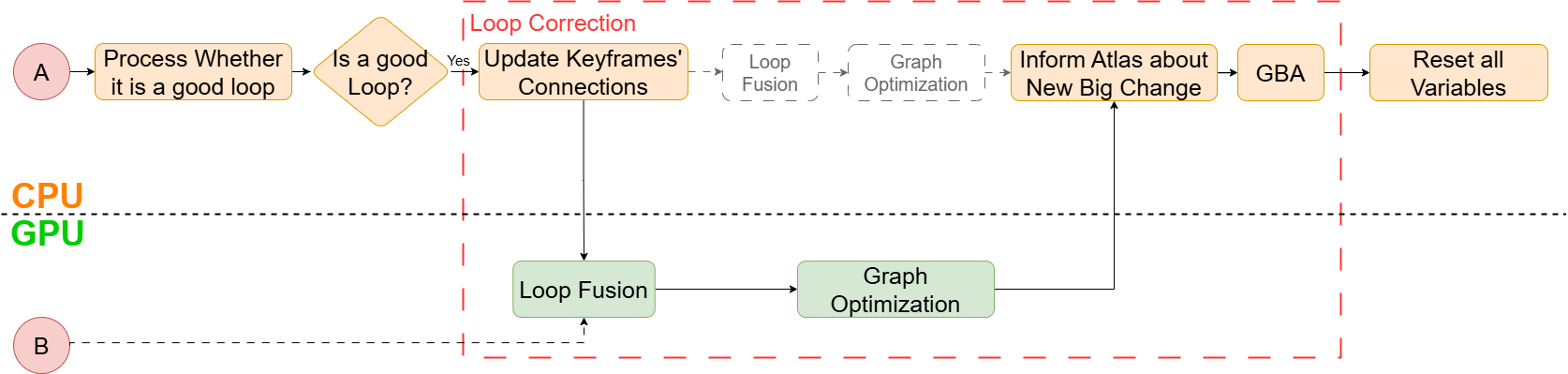}
        \caption{Loop Correction components.
        }
        \label{fig:fastLoop_breakdown_b}
    \end{subfigure}

    \caption{
    Overview of the \SYS{} loop closing process. Top: CPU modules; bottom: GPU modules, with arrows showing data flow. Keyframes reside in GPU memory. White blocks mark components parallelized by \SYS{} but were originally executed on the CPU.
    }
    \label{fig:fastLoop_breakdown}
    \vspace{-15pt}
\end{figure*}

\section{Design} \label{sec:design}

\subsection{System Overview}
\label{design:Overview}


\autoref{fig:fastLoop_breakdown} presents the loop closing architecture of \SYS{}, illustrating how responsibilities are divided between the CPU and GPU and how we parallelize the computations.
White dotted blocks denote components that we parallelize, components that are originally sequential 
but are now parallelized on the GPU in FastLoop.
As illustrated, the loop closing stage is triggered whenever a new keyframe is inserted into the map. Similar to the keyframe storage strategy adopted in TurboMap~\cite{hosseininejad2026turbomapgpuacceleratedlocalmapping}, we first
transfer each newly created keyframe to GPU-resident KeyFrame Storage to avoid redundant
data transfers between CPU memory and GPU memory.

\SYS{} transfers to the GPU only the immutable portion of each keyframe and map point, data that does not change once created, and therefore requires no CPU synchronization once resident on the GPU. Mutable data, which can be updated during tracking, local mapping, or loop closing, remains CPU-resident and continues to be protected by the same synchronization primitives the pipeline already uses across its threads.
\SYS{} therefore changes only the computation over immutable data.

\autoref{fig:loop_correction_time_breakdown} shows that Loop Fusion and Graph Optimization dominate Loop Correction's computation time.
Motivated by this bottleneck, we analyze the computation and data flow of each component in the loop closing to identify parallelization opportunities. For Graph Optimization, we adopt Graphite~\cite{gopinath2026graphitegpuacceleratedmixedprecisiongraph}, a GPU-accelerated graph optimization framework, described later. For the remaining components, we find that Projection Search and Loop Fusion each exhibit inherent task- and data-level parallelism, discussed in detail below, and redesign both in \SYS{}.

\subsection{Parallelization Strategies}
\label{design:parallelization_strategies}

Loop closure corrects every frame in the loop to adhere to the visual odometry constraints. Empirically, we observe that this involves either repeatedly performing the same set of operations on several parts of the map, or performing multiple operations on the same data. Task-level and data-level parallelism are well-established techniques for exploiting exactly these patterns. Our contribution is
in identifying which parts of the loop closing pipeline are amenable to each type of parallelism, and in redesigning their computation and data flow to expose these opportunities while reducing data movement between the CPU and GPU.

Task-level parallelism executes independent computational tasks concurrently. We identify tasks within loop closure that (1) use the same type of input data, (2) have no data dependencies, and (3) do not immediately consume each other's outputs downstream, and execute them concurrently on the GPU. Since these tasks share the same input type, we transfer that data to the GPU only once and reuse it across all of them,
avoiding repeated CPU--GPU transfers.

Data-level parallelism applies the same operation to many independent data elements simultaneously. We identify operations within loop closure that (1) apply an identical computation across different map points or keyframes and (2) have no dependencies among those elements, and execute them in parallel on the GPU. For these operations, we reduce back-and-forth CPU--GPU transfers and redundant kernel launches, reducing
overall execution time.

In the rest of this section, we discuss how we apply each approach in detail and identify the components that meet the criteria and have been parallelized accordingly.

\subsection{Task Parallelism}
\label{design:tasK_parallelism}

As discussed in \autoref{design:Overview},
Projection Search presents an opportunity for task-level parallelism. Within this stage, we identify two operations,
PS2a and PS2b, shown in \autoref{fig:fastLoop_breakdown_a}, that in the original sequential pipeline are separated by Sim(3) optimization. We observe that these two tasks meet the criteria for task parallelism described above, because (1) the two tasks operate on almost the same set of map points and on the same keyframe (current keyframe in the pipeline); (2) their internal procedures differ only slightly, and they have no data dependency even when executed multiple times; and (3) their outputs are not required at the same time. Motivated by this observation, we redesign the data flow and change the order of execution, running some computations earlier than in the original design. This spreads the workload over time and reduces execution time when a loop is detected.

We further find that whether PS2a and PS2b can be safely parallelized depends on the camera model. For stereo fisheye configurations, the original implementation already computes both projection searches from the same pre-optimization pose within a single merged routine; in this case, PS2a and PS2b have no data dependency, and \SYS{} executes them concurrently on the GPU without modification to correctness.


For standard pinhole stereo, however, the original pipeline computes PS2b using a pose refined by the interim Sim(3) optimization, so PS2b genuinely depends on PS2a's output. To exploit parallelism here while preserving correctness, \SYS{} executes PS2b speculatively, concurrently with PS2a, using the pre-optimization pose. Once Sim(3) optimization completes, we reuse the pipeline's existing acceptance criterion to check whether the refined pose differs meaningfully from the speculative estimate. If it passes, the speculative result is accepted directly; otherwise, PS2b is re-executed with the refined pose, exactly reproducing the original sequential computation. \SYS{} thus never produces a different result from unmodified ORB-SLAM3; only the timing of PS2b's computation changes when speculation is validated as safe.


\subsection{Data Parallelism}
\label{design:data_parallelism}



We analyze the computation and data flow of each component in loop closing and identify three that repeat the same operation independently across a large number of map points or keyframes, the pattern that data-level parallelism exploits.
\subsubsection{Single Projection Search}
\label{design:single_projection_search}

The first component is Projection Search (PS1, as shown in \autoref{fig:fastLoop_breakdown_a}), performed during Detect and Refine Sim3 from the Last Keyframe to find correspondences between local map points and the candidate keyframe's features. We analyze its computational profile and data access pattern to determine whether it is amenable to parallelization. This task (1) is computationally expensive, evaluating a projection and descriptor match for every local map point against the candidate keyframe, and (2) operates over a large set of independent map points, since each point's projection and matching outcome depends only on its own position and descriptor. These two properties, high per-invocation cost and independence across a large dataset, are precisely the conditions under which data-level parallelism yields substantial benefit. We therefore decouple Projection Search from the rest of the component and parallelize it on the GPU. After transferring the current keyframe and map point data, we launch a kernel in which each thread handles the projection and matching for a single map point. Results are then transferred back to the CPU in a single batch, where final correspondences are identified and validated.

\subsubsection{Triple Projection Search}
\label{design:3D_projection_search}
The second component is the Projection Search tasks labeled PS3a-PS3c during Region Detection from Last Keyframe, which are invoked for three consecutive keyframes in the original sequential design. The reason for the three invocations is to mitigate false positives. Instead of confirming region detection with a single keyframe, the original design uses three consecutive keyframes. Although this is good for false positives, it can delay or even prevent loop detection. We observe that these three Projection Search operations perform the same computation independently on three different keyframes, making them well-suited for data-level parallelism. To exploit this, we process them concurrently on the GPU: we transfer all the data required for the three operations to the GPU once and launch a single kernel to execute the computations simultaneously, as illustrated in \autoref{fig:fastLoop_breakdown_a}. The resulting matches are then merged to determine whether a valid loop has been detected.

\subsubsection{Loop Fusion}
\label{design:loop_fusion}

The third component is loop fusion, which detects and fuses duplicate map points during loop correction, introducing new edges into the covisibility and essential graphs. A map point is fused if it (1) has a descriptor similar to one observed in the current keyframe and (2) lies in close spatial proximity. This stage is computationally intensive, iterating over multiple connected keyframes and evaluating their map points against these criteria. However, computations across keyframes are mutually independent, making the process well-suited for data-level parallelism. We exploit this by decoupling and executing the per-keyframe evaluations concurrently on the GPU, reusing GPU-resident keyframe data to reduce transfer overhead (\autoref{fig:fastLoop_breakdown_b}). As the number of connected keyframes and graph size grow, speedup becomes more pronounced, since computation increasingly dominates transfer overhead.

\subsection{GPU Memory and Data Transfer Optimization}
\label{design:GPU_memory_and_data_transfer Optimization}

Another practical challenge is the high cost of data transfer between CPU and GPU memory. To mitigate this overhead, we apply several optimizations. First, as explained earlier, we adopt GPU-based keyframe storage~\cite{hosseininejad2026turbomapgpuacceleratedlocalmapping}.
This design allows subsequent components to access the required data directly on the GPU, eliminating redundant transfers and distributing the workload over time. 
Second, we carefully analyze the data required during GPU execution and transfer only the necessary keyframe and map point information. Before transferring data to the GPU, we preprocess and organize it. Specifically, we collect only the data elements from Keyframe and map points that are actually used in the GPU kernel into a lightweight wrapper structure. Therefore, instead of copying the full keyframe and map point objects, we copy this smaller structure to the GPU to reduce the transfer volume and improve efficiency.
Third, as illustrated in \autoref{fig:fastLoop_breakdown_a}, multiple Projection Search components are parallelized. Initializing GPU memory separately for each would introduce significant overhead and negate the benefits of our optimization strategy. Since these Projection Search components require nearly identical memory layouts, we allocate the GPU memory once during the execution of the first component and retain it in GPU memory for reuse by the subsequent components. The memory is released only when the system terminates. This approach avoids redundant memory allocations and reduces overhead. Finally, we employ pinned (page-locked) memory to accelerate host-to-device data transfers. Unlike pageable memory, pinned memory prevents the OS from swapping the allocated region, allowing the GPU to access it directly through DMA (Direct Memory Access). This eliminates the need for an additional intermediate copy performed by the CUDA driver when transferring data from pageable memory, thereby reducing transfer latency. As a result, we reduce data transfer time and improve the overall efficiency of GPU-CPU interaction.

\subsection{Graph Optimization}
\label{design:graph_optimization}

The next step is an essential (pose) graph optimization to propagate the loop correction to the rest of the map, as shown in \autoref{fig:fastLoop_breakdown_b}.
To move this process to the GPU, we rewrite the original pose graph optimization and replace the optimization library, g$^2$o~\cite{Kümmerle2011}, with a GPU-accelerated alternative, Graphite~\cite{gopinath2026graphitegpuacceleratedmixedprecisiongraph}. That is, we implement our own optimizable poses and constraints, construct the pose graph, and optimize it on the GPU with the Levenberg-Marquardt algorithm. Within this Levenberg--Marquardt optimization loop, every step except the final linear solve is executed on the GPU: linearization (Jacobian assembly via automatic differentiation), error/residual computation, and the variable update step are all performed as GPU kernels across iterations. For consistency with the original implementation, we extend Graphite by adding support for direct solvers, including CPU-based Eigen LDLT and GPU-based cuDSS, for which we compute and store the sparse Hessian on the GPU. We select CPU-based Eigen LDLT for the linear solver because it performs better than GPU-based cuDSS for smaller problem sizes (\autoref{sec:evaluation}). To compute Jacobian matrices, we use automatic differentiation, which provides more exact derivatives than the numerical differentiation used previously.

\begin{table*}[ht]
  \centering
  \begin{subtable}{\textwidth}
    \centering
    \renewcommand{\arraystretch}{1.8}
\vspace{10pt}
\centering
\resizebox{1\linewidth}{!}{
\setlength{\tabcolsep}{4pt}
\begin{tabular}{|c|
*{2}{>{\centering\arraybackslash}p{1.9cm}|}p{0.7cm}|
*{2}{>{\centering\arraybackslash}p{1.0cm}|}p{0.7cm}|
*{2}{>{\centering\arraybackslash}p{0.7cm}|}p{0.7cm}|
*{2}{>{\centering\arraybackslash}p{1.0cm}|}p{0.7cm}|
*{2}{>{\centering\arraybackslash}p{1.0cm}|}p{0.7cm}|
*{2}{>{\centering\arraybackslash}p{0.8cm}|}}
\hline
\multirow{2}{*}{\textbf{Sequence}} &
\multicolumn{2}{c|}{\textbf{Total Loop Closing}} & \multicolumn{1}{c|}{\textbf{Speed}} &
\multicolumn{2}{c|}{\textbf{Region Detection}} & \multicolumn{1}{c|}{\textbf{Speed}} &
\multicolumn{2}{c|}{\textbf{Loop Fusion}} & \multicolumn{1}{c|}{\textbf{Speed}} &
\multicolumn{2}{c|}{\textbf{Graph Optimization}} & \multicolumn{1}{c|}{\textbf{Speed}} &
\multicolumn{2}{c|}{\textbf{Loop Correction}} & \multicolumn{1}{c|}{\textbf{Speed}} &
\multicolumn{2}{c|}{\textbf{ATE}} \\
\cline{2-3} \cline{5-6} \cline{8-9} \cline{11-12} \cline{14-15} \cline{17-18}
 & \textbf{Original} & \textbf{\SYS{}} & \multicolumn{1}{c|}{\textbf{Up}} 
 & \textbf{Org} & \textbf{FL} & \multicolumn{1}{c|}{\textbf{Up}} 
 & \textbf{Org} & \textbf{FL} & \multicolumn{1}{c|}{\textbf{Up}} 
 & \textbf{Org} & \textbf{FL} & \multicolumn{1}{c|}{\textbf{Up}}
 & \textbf{Org} & \textbf{FL} & \multicolumn{1}{c|}{\textbf{Up}}
 & \textbf{Org} & \textbf{FL} \\
\hline
\Xhline{4\arrayrulewidth}
\textbf{EuRoC Avg}
& 153.8 & 114.0 & \textcolor{Green}{\textbf{\hspace{+0.09cm}{1.3x}}} 
& 34.4 & 31.1 & \textcolor{Green}{\textbf{\hspace{+0.09cm}{1.1x}}} 
& 23.6 & 8.4 & \textcolor{Green}{\textbf{\hspace{+0.09cm}{2.8x}}} 
& 60.5 & 47.2 & \textcolor{Green}{\textbf{\hspace{+0.09cm}{1.3x}}} 
& 119.4 & 83.9 & \textcolor{Green}{\textbf{\hspace{+0.09cm}{1.4x}}} 
& 0.044 & 0.030 \\
\hline
\textbf{V102} 
& $155.9 \pm 29.4$ & $124.4 \pm 3.9$ & \textcolor{Green}{\textbf{\hspace{+0.09cm}{1.3x}}} 
& $39.2$ & $31.1$ & \textcolor{Green}{\textbf{\hspace{+0.09cm}{1.3x}}} 
& $17.9$ & $10.2$ & \textcolor{Green}{\textbf{\hspace{+0.09cm}{1.8x}}} 
& $52.8$ & $49.3$ & \textcolor{Green}{\textbf{\hspace{+0.09cm}{1.1x}}} 
& $116.6$ & $92.2$ & \textcolor{Green}{\textbf{\hspace{+0.09cm}{1.3x}}} 
& $0.019$ & $0.018$ \\
\hline
\textbf{V103} 
& $71.9 \pm 0.8$ & $74.3 \pm 0.5$ & \textcolor{Black}{\textbf{\hspace{+0.09cm}{1.0x}}} 
& $30.9$ & $32.2$ & \textcolor{Black}{\textbf{\hspace{+0.09cm}{1.0x}}} 
& $14.5$ & $5.1$ & \textcolor{Green}{\textbf{\hspace{+0.09cm}{2.8x}}} 
& $18.6$ & $28.1$ & \textcolor{Red}{\textbf{\hspace{+0.09cm}{0.7x}}} 
& $40.8$ & $41.9$ & \textcolor{Black}{\textbf{\hspace{+0.09cm}{1.0x}}} 
& $0.024$ & $0.024$ \\
\hline
\textbf{V202} 
& $171.8 \pm 81.3$ & $111.0 \pm 40.9$ & \textcolor{Green}{\textbf{\hspace{+0.09cm}{1.5x}}} 
& $31.8$ & $25.8$ & \textcolor{Green}{\textbf{\hspace{+0.09cm}{1.2x}}} 
& $23.7$ & $7.7$ & \textcolor{Green}{\textbf{\hspace{+0.09cm}{3.1x}}} 
& $75.1$ & $45.4$ & \textcolor{Green}{\textbf{\hspace{+0.09cm}{1.7x}}} 
& $139.5$ & $84.4$ & \textcolor{Green}{\textbf{\hspace{+0.09cm}{1.7x}}} 
& $0.020$ & $0.015$ \\
\hline
\textbf{V203} 
& $215.6 \pm 34.0$ & $146.2 \pm 32.7$ & \textcolor{Green}{\textbf{\hspace{+0.09cm}{1.5x}}} 
& $35.7$ & $35.3$ & \textcolor{Black}{\textbf{\hspace{+0.09cm}{1.0x}}} 
& $38.5$ & $10.8$ & \textcolor{Green}{\textbf{\hspace{+0.09cm}{3.6x}}} 
& $95.6$ & $66.1$ & \textcolor{Green}{\textbf{\hspace{+0.09cm}{1.4x}}} 
& $180.6$ & $117.0$ & \textcolor{Green}{\textbf{\hspace{+0.09cm}{1.5x}}} 
& $0.112$ & $0.062$ \\
\hline
\Xhline{4\arrayrulewidth}
\textbf{TUM-VI Avg}
& 1196.5 & 504.4 & \textcolor{Green}{\textbf{\hspace{+0.09cm}{2.4x}}} 
& 29.1 & 26.8 & \textcolor{Green}{\textbf{\hspace{+0.09cm}{1.1x}}} 
& 78.7 & 14.8 & \textcolor{Green}{\textbf{\hspace{+0.09cm}{5.3x}}} 
& 927.8 & 351.2 & \textcolor{Green}{\textbf{\hspace{+0.09cm}{2.6x}}} 
& 1166.6 & 477.4 & \textcolor{Green}{\textbf{\hspace{+0.09cm}{2.4x}}} 
& 5.900 & 4.700 \\
\hline
\textbf{room3} 
& $136.1 \pm 17.9$ & $110.8 \pm 21.4$ & \textcolor{Green}{\textbf{\hspace{+0.09cm}{1.2x}}} 
& $26.9$ & $24.1$ & \textcolor{Green}{\textbf{\hspace{+0.09cm}{1.1x}}} 
& $31.8$ & $9.0$ & \textcolor{Green}{\textbf{\hspace{+0.09cm}{3.5x}}} 
& $51.0$ & $54.3$ & \textcolor{Red}{\textbf{\hspace{+0.09cm}{0.9x}}} 
& $110.1$ & $86.9$ & \textcolor{Green}{\textbf{\hspace{+0.09cm}{1.3x}}} 
& $0.009$ & $0.008$ \\
\hline
\textbf{room4} 
& $308.9 \pm 45.4$ & $163.6 \pm 12.1$ & \textcolor{Green}{\textbf{\hspace{+0.09cm}{1.9x}}} 
& $33.8$ & $32.3$ & \textcolor{Black}{\textbf{\hspace{+0.09cm}{1.0x}}} 
& $88.6$ & $14.3$ & \textcolor{Green}{\textbf{\hspace{+0.09cm}{6.2x}}} 
& $72.1$ & $64.4$ & \textcolor{Green}{\textbf{\hspace{+0.09cm}{1.1x}}} 
& $277.0$ & $130.7$ & \textcolor{Green}{\textbf{\hspace{+0.09cm}{2.1x}}} 
& $0.007$ & $0.007$ \\
\hline
\textbf{corridor1} 
& $1106.8 \pm 93.2$ & $359.8 \pm 6.8$ & \textcolor{Green}{\textbf{\hspace{+0.09cm}{3.1x}}} 
& $29.8$ & $27.2$ & \textcolor{Green}{\textbf{\hspace{+0.09cm}{1.1x}}} 
& $18.3$ & $8.9$ & \textcolor{Green}{\textbf{\hspace{+0.09cm}{2.1x}}} 
& $956.9$ & $289.2$ & \textcolor{Green}{\textbf{\hspace{+0.09cm}{3.3x}}} 
& $1076.9$ & $333.8$ & \textcolor{Green}{\textbf{\hspace{+0.09cm}{3.2x}}} 
& $0.035$ & $0.058$ \\
\hline
\textbf{magistrale1} 
& $1083.9 \pm 326.6$ & $368.7 \pm 124.6$ & \textcolor{Green}{\textbf{\hspace{+0.09cm}{2.9x}}} 
& $23.0$ & $22.1$ & \textcolor{Black}{\textbf{\hspace{+0.09cm}{1.0x}}} 
& $57.8$ & $8.8$ & \textcolor{Green}{\textbf{\hspace{+0.09cm}{6.6x}}} 
& $917.3$ & $295.2$ & \textcolor{Green}{\textbf{\hspace{+0.09cm}{3.1x}}} 
& $1058.4$ & $347.7$ & \textcolor{Green}{\textbf{\hspace{+0.09cm}{3.0x}}} 
& $0.775$ & $0.562$ \\
\hline
\textbf{magistrale2} 
& $1072.6 \pm 80.8$ & $420.7 \pm 41.0$ & \textcolor{Green}{\textbf{\hspace{+0.09cm}{2.5x}}} 
& $29.3$ & $23.2$ & \textcolor{Green}{\textbf{\hspace{+0.09cm}{1.3x}}} 
& $140.2$ & $16.6$ & \textcolor{Green}{\textbf{\hspace{+0.09cm}{8.4x}}} 
& $691.9$ & $252.0$ & \textcolor{Green}{\textbf{\hspace{+0.09cm}{2.7x}}} 
& $1041.3$ & $397.1$ & \textcolor{Green}{\textbf{\hspace{+0.09cm}{2.6x}}} 
& $0.577$ & $0.641$ \\
\hline
\textbf{outdoors5} 
& $2166.8 \pm 254.6$ & $897.8 \pm 173.9$ & \textcolor{Green}{\textbf{\hspace{+0.09cm}{2.4x}}} 
& $30.3$ & $28.3$ & \textcolor{Green}{\textbf{\hspace{+0.09cm}{1.1x}}} 
& $153.5$ & $26.8$ & \textcolor{Green}{\textbf{\hspace{+0.09cm}{5.7x}}} 
& $1495.7$ & $461.6$ & \textcolor{Green}{\textbf{\hspace{+0.09cm}{3.2x}}} 
& $2137.5$ & $867.2$ & \textcolor{Green}{\textbf{\hspace{+0.09cm}{2.5x}}} 
& $33.700$ & $27.758$ \\
\hline
\textbf{outdoors7} 
& $2500.0 \pm 1016.9$ & $1209.5 \pm 817.7$ & \textcolor{Green}{\textbf{\hspace{+0.09cm}{2.1x}}} 
& $30.8$ & $30.5$ & \textcolor{Black}{\textbf{\hspace{+0.09cm}{1.0x}}} 
& $60.7$ & $19.1$ & \textcolor{Green}{\textbf{\hspace{+0.09cm}{3.2x}}} 
& $2309.3$ & $1041.7$ & \textcolor{Green}{\textbf{\hspace{+0.09cm}{2.2x}}} 
& $2465.3$ & $1178.2$ & \textcolor{Green}{\textbf{\hspace{+0.09cm}{2.1x}}} 
& $5.900$ & $3.810$ \\
\hline

\end{tabular}
}
\label{tab:jetson_runtime_comparison}


    \caption{Jetson Results}
    \label{tab:jetson}
  \end{subtable}
  
  \begin{subtable}{\textwidth}
    \centering
    \renewcommand{\arraystretch}{1.8}
\vspace{10pt}
\centering
\resizebox{1\linewidth}{!}{
\setlength{\tabcolsep}{4pt}
\begin{tabular}{|c|
*{2}{>{\centering\arraybackslash}p{1.9cm}|}p{0.7cm}|
*{2}{>{\centering\arraybackslash}p{1.0cm}|}p{0.7cm}|
*{2}{>{\centering\arraybackslash}p{1.0cm}|}p{0.7cm}|
*{2}{>{\centering\arraybackslash}p{1.0cm}|}p{0.7cm}|
*{2}{>{\centering\arraybackslash}p{1.0cm}|}p{0.7cm}|
*{2}{>{\centering\arraybackslash}p{0.8cm}|}}
\hline
\multirow{2}{*}{\textbf{Sequence}} &
\multicolumn{2}{c|}{\textbf{Total Loop Closing}} & \multicolumn{1}{c|}{\textbf{Speed}} &
\multicolumn{2}{c|}{\textbf{Region Detection}} & \multicolumn{1}{c|}{\textbf{Speed}} &
\multicolumn{2}{c|}{\textbf{Loop Fusion}} & \multicolumn{1}{c|}{\textbf{Speed}} &
\multicolumn{2}{c|}{\textbf{Graph Optimization}} & \multicolumn{1}{c|}{\textbf{Speed}} &
\multicolumn{2}{c|}{\textbf{Loop Correction}} & \multicolumn{1}{c|}{\textbf{Speed}} &
\multicolumn{2}{c|}{\textbf{ATE}} \\
\cline{2-3} \cline{5-6} \cline{8-9} \cline{11-12} \cline{14-15} \cline{17-18}
 & \textbf{Original} & \textbf{\SYS{}} & \multicolumn{1}{c|}{\textbf{Up}} 
 & \textbf{Org} & \textbf{FL} & \multicolumn{1}{c|}{\textbf{Up}} 
 & \textbf{Org} & \textbf{FL} & \multicolumn{1}{c|}{\textbf{Up}} 
 & \textbf{Org} & \textbf{FL} & \multicolumn{1}{c|}{\textbf{Up}}
 & \textbf{Org} & \textbf{FL} & \multicolumn{1}{c|}{\textbf{Up}}
 & \textbf{Org} & \textbf{FL} \\
\Xhline{4\arrayrulewidth}
\textbf{EuRoC Avg}
& 56.9 & 39.8 & \textcolor{Green}{\textbf{\hspace{+0.09cm}{1.4×}}}
& 13.7 & 13.3 & \textcolor{Black}{\textbf{\hspace{+0.09cm}{1.0×}}}
& 7.8 & 3.4 & \textcolor{Green}{\textbf{\hspace{+0.09cm}{2.3×}}}
& 24.6 & 16.1 & \textcolor{Green}{\textbf{\hspace{+0.09cm}{1.5×}}}
& 43.9 & 29.0 & \textcolor{Green}{\textbf{\hspace{+0.09cm}{1.5×}}}
& 0.033 & 0.025 \\
\hline
\textbf{V102}
& $60.5 \pm 4.9$ & $45.0 \pm 4.5$ & \textcolor{Green}{\textbf{\hspace{+0.09cm}{1.3x}}}
& $22.2$ & $19.8$ & \textcolor{Green}{\textbf{\hspace{+0.09cm}{1.1x}}}
& $7.5$ & $3.6$ & \textcolor{Green}{\textbf{\hspace{+0.09cm}{2.1x}}}
& $21.8$ & $18.0$ & \textcolor{Green}{\textbf{\hspace{+0.09cm}{1.2x}}}
& $41.3$ & $30.4$ & \textcolor{Green}{\textbf{\hspace{+0.09cm}{1.4x}}}
& $0.018$ & $0.018$ \\
\hline
\textbf{V103}
& $23.6 \pm 1.7$ & $29.6 \pm 1.3$ & \textcolor{Red}{\textbf{\hspace{+0.09cm}{0.8x}}}
& $8.8$ & $10.0$ & \textcolor{Red}{\textbf{\hspace{+0.09cm}{0.9x}}}
& $2.0$ & $2.5$ & \textcolor{Red}{\textbf{\hspace{+0.09cm}{0.8x}}}
& $8.4$ & $11.0$ & \textcolor{Red}{\textbf{\hspace{+0.09cm}{0.8x}}}
& $14.5$ & $19.8$ & \textcolor{Red}{\textbf{\hspace{+0.09cm}{0.7x}}}
& $0.024$ & $0.025$ \\
\hline
\textbf{V202}
& $52.9 \pm 13.1$ & $34.3 \pm 1.5$ & \textcolor{Green}{\textbf{\hspace{+0.09cm}{1.5x}}}
& $10.8$ & $10.2$ & \textcolor{Green}{\textbf{\hspace{+0.09cm}{1.1x}}}
& $8.1$ & $3.1$ & \textcolor{Green}{\textbf{\hspace{+0.09cm}{2.6x}}}
& $22.5$ & $11.9$ & \textcolor{Green}{\textbf{\hspace{+0.09cm}{1.9x}}}
& $42.1$ & $24.0$ & \textcolor{Green}{\textbf{\hspace{+0.09cm}{1.8x}}}
& $0.026$ & $0.012$ \\
\hline
\textbf{V203}
& $90.6 \pm 11.5$ & $50.4 \pm 3.8$ & \textcolor{Green}{\textbf{\hspace{+0.09cm}{1.8x}}}
& $13.1$ & $13.0$ & \textcolor{Black}{\textbf{\hspace{+0.09cm}{1.0x}}}
& $13.7$ & $4.5$ & \textcolor{Green}{\textbf{\hspace{+0.09cm}{3.0x}}}
& $45.7$ & $23.5$ & \textcolor{Green}{\textbf{\hspace{+0.09cm}{1.9x}}}
& $77.7$ & $41.8$ & \textcolor{Green}{\textbf{\hspace{+0.09cm}{1.9x}}}
& $0.062$ & $0.044$ \\
\hline
\Xhline{4\arrayrulewidth}
\textbf{TUM-VI Avg}
& 485.3 & 163.7 & \textcolor{Green}{\textbf{\hspace{+0.09cm}{3.0×}}}
& 11.4 & 10.1 & \textcolor{Green}{\textbf{\hspace{+0.09cm}{1.1×}}}
& 23.2 & 7.0 & \textcolor{Green}{\textbf{\hspace{+0.09cm}{3.3×}}}
& 392.7 & 99.1 & \textcolor{Green}{\textbf{\hspace{+0.09cm}{4.0×}}}
& 473.7 & 152.9 & \textcolor{Green}{\textbf{\hspace{+0.09cm}{3.1×}}}
& 3.128 & 3.347 \\
\hline
\textbf{room3}
& $47.8 \pm 8.1$ & $36.6 \pm 2.9$ & \textcolor{Green}{\textbf{\hspace{+0.09cm}{1.3x}}}
& $10.3$ & $9.2$ & \textcolor{Green}{\textbf{\hspace{+0.09cm}{1.1x}}}
& $13.2$ & $4.4$ & \textcolor{Green}{\textbf{\hspace{+0.09cm}{3.0x}}}
& $14.7$ & $13.6$ & \textcolor{Green}{\textbf{\hspace{+0.09cm}{1.1x}}}
& $37.4$ & $26.8$ & \textcolor{Green}{\textbf{\hspace{+0.09cm}{1.4x}}}
& $0.008$ & $0.009$ \\
\hline
\textbf{room4}
& $72.4 \pm 11.4$ & $59.6 \pm 8.9$ & \textcolor{Green}{\textbf{\hspace{+0.09cm}{1.2x}}}
& $12.3$ & $11.0$ & \textcolor{Green}{\textbf{\hspace{+0.09cm}{1.1x}}}
& $13.6$ & $7.0$ & \textcolor{Green}{\textbf{\hspace{+0.09cm}{1.9x}}}
& $26.6$ & $16.3$ & \textcolor{Green}{\textbf{\hspace{+0.09cm}{1.6x}}}
& $59.9$ & $47.6$ & \textcolor{Green}{\textbf{\hspace{+0.09cm}{1.3x}}}
& $0.007$ & $0.006$ \\
\hline
\textbf{corridor1}
& $558.7 \pm 55.1$ & $152.7 \pm 4.3$ & \textcolor{Green}{\textbf{\hspace{+0.09cm}{3.7x}}}
& $13.7$ & $12.0$ & \textcolor{Green}{\textbf{\hspace{+0.09cm}{1.1x}}}
& $14.8$ & $5.0$ & \textcolor{Green}{\textbf{\hspace{+0.09cm}{3.0x}}}
& $499.2$ & $117.0$ & \textcolor{Green}{\textbf{\hspace{+0.09cm}{4.3x}}}
& $545.1$ & $139.8$ & \textcolor{Green}{\textbf{\hspace{+0.09cm}{3.9x}}}
& $0.030$ & $0.030$ \\
\hline
\textbf{magistrale1}
& $471.0 \pm 149.1$ & $148.4 \pm 55.2$ & \textcolor{Green}{\textbf{\hspace{+0.09cm}{3.2x}}}
& $8.7$ & $8.6$ & \textcolor{Black}{\textbf{\hspace{+0.09cm}{1.0x}}}
& $9.4$ & $4.8$ & \textcolor{Green}{\textbf{\hspace{+0.09cm}{2.0x}}}
& $429.3$ & $108.9$ & \textcolor{Green}{\textbf{\hspace{+0.09cm}{3.9x}}}
& $460.8$ & $139.4$ & \textcolor{Green}{\textbf{\hspace{+0.09cm}{3.3x}}}
& $0.596$ & $0.435$ \\
\hline
\textbf{magistrale2}
& $366.1 \pm 63.4$ & $133.0 \pm 10.4$ & \textcolor{Green}{\textbf{\hspace{+0.09cm}{2.8x}}}
& $9.8$ & $8.6$ & \textcolor{Green}{\textbf{\hspace{+0.09cm}{1.1x}}}
& $36.4$ & $8.3$ & \textcolor{Green}{\textbf{\hspace{+0.09cm}{4.4x}}}
& $242.9$ & $73.7$ & \textcolor{Green}{\textbf{\hspace{+0.09cm}{3.3x}}}
& $356.4$ & $123.8$ & \textcolor{Green}{\textbf{\hspace{+0.09cm}{2.9x}}}
& $0.795$ & $0.664$ \\
\hline
\textbf{outdoors5}
& $949.7 \pm 183.9$ & $350.1 \pm 53.0$ & \textcolor{Green}{\textbf{\hspace{+0.09cm}{2.7x}}}
& $11.7$ & $9.8$ & \textcolor{Green}{\textbf{\hspace{+0.09cm}{1.2x}}}
& $49.3$ & $13.6$ & \textcolor{Green}{\textbf{\hspace{+0.09cm}{3.6x}}}
& $689.8$ & $147.5$ & \textcolor{Green}{\textbf{\hspace{+0.09cm}{4.7x}}}
& $937.3$ & $339.5$ & \textcolor{Green}{\textbf{\hspace{+0.09cm}{2.8x}}}
& $16.806$ & $16.064$ \\
\hline
\textbf{outdoors7}
& $930.994 \pm 447.3$ & $265.7 \pm 84.2$ & \textcolor{Green}{\textbf{\hspace{+0.09cm}{3.5x}}}
& $13.3$ & $11.2$ & \textcolor{Green}{\textbf{\hspace{+0.09cm}{1.2x}}}
& $25.4$ & $6.1$ & \textcolor{Green}{\textbf{\hspace{+0.09cm}{4.2x}}}
& $846.1$ & $216.3$ & \textcolor{Green}{\textbf{\hspace{+0.09cm}{3.9x}}}
& $919.0$ & $253.5$ & \textcolor{Green}{\textbf{\hspace{+0.09cm}{3.6x}}}
& $3.655$ & $6.223$ \\
\hline
\end{tabular}
}
\label{tab:desktop_runtime_comparison}

    \caption{Desktop Results}
    \label{tab:desktop}
  \end{subtable}

  \caption{
  Runtime (mean ± std, ms), speed-up, and ATE RMSE (m) for \SYS{} vs.\ ORB-SLAM3 on Desktop and Jetson, using \EUROC{} and \TUMVI{}. Runtimes include H2D/D2H transfer and synchronization overhead.
  }
  \label{tab:combined_results}
  \vspace{-15pt}
\end{table*}

    

    
    
    
    

\section{Evaluation} \label{sec:evaluation}


Loop closure events are rare, but when they occur, a single occurrence can take orders of magnitude longer than a normal iteration, creating latency spikes that can stall tracking and local mapping. Our evaluation therefore focuses not only on overall runtime, but on how effectively \SYS{} reduces these per-occurrence latency spikes and their impact on system responsiveness.

\subsection{Experimental Setup}
\label{eval:setup}


We evaluate \SYS{} against the original ORB-SLAM3 on two hardware platforms: (1) a desktop system with an 8-core Intel i9-11900 CPU at 2.5 GHz, an NVIDIA GeForce RTX 3060 Ti GPU with 4,864 CUDA cores, and 32 GB RAM; and (2) an embedded NVIDIA Jetson Orin Nano Developer Kit with a 6-core Arm Cortex-A78AE CPU running at up to 1.7 GHz (MAXN SUPER mode), a 1024-core Ampere GPU with 32 Tensor Cores, and 8 GB LPDDR5 RAM. To ensure fair and repeatable comparisons, both CPU and GPU frequencies are fixed at their respective maximum operating values throughout all experiments.


For evaluation, we use a combination of sequences from the \TUMVI{} and \EUROC{} datasets that provide a loop closure of varying lengths with the stereo-inertial configuration. We run each sequence 5 times and report the average results across these runs to account for variability and ensure fair comparison.

\begin{figure}[t]
    \centering
    \includegraphics[width=0.65\linewidth]{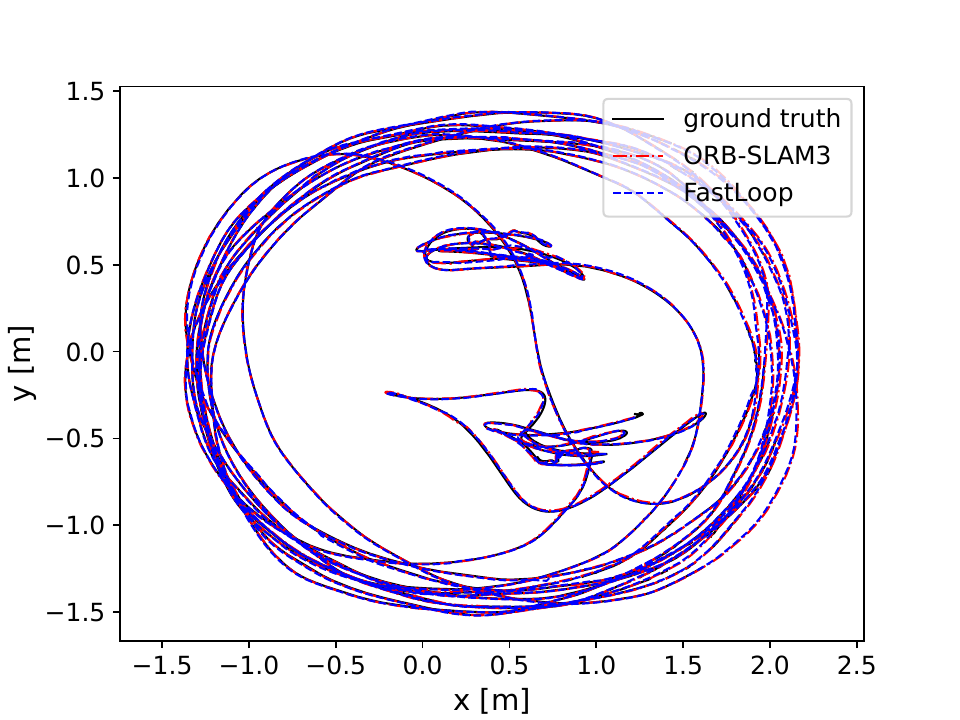}
    \caption{Trajectory comparison between \SYS{} and ORB-SLAM3 with the ground truth in Room3.}
    \label{fig:room3_trajectory_comparison}
    \vspace{-20pt}
\end{figure}

\subsection{Overall Performance}
\label{eval:overall_performance}



\begin{figure}[!t]
    \centering
    
    \begin{subfigure}[t]{0.8\linewidth}
        \centering
        \includegraphics[width=\linewidth]{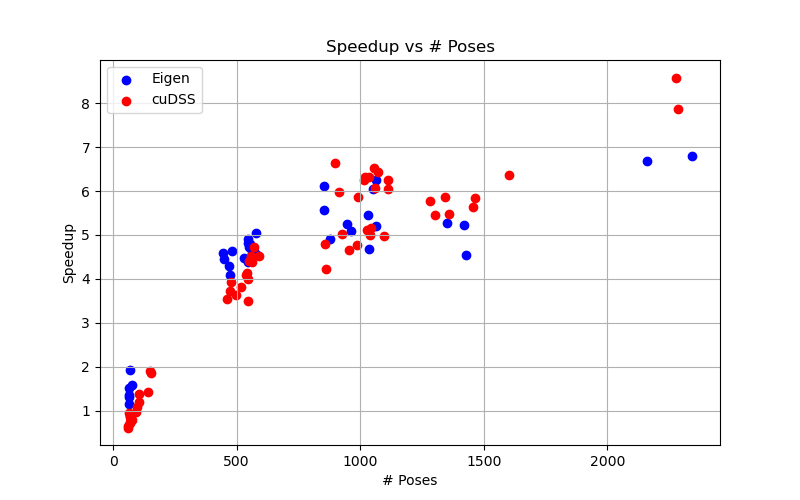}
        \caption{Speedup vs \# of Poses}
        \label{fig:graph_optimization_speedup_a}
    \end{subfigure}
    
    
    \begin{subfigure}[t]{0.8\linewidth}
        \centering
        \includegraphics[width=\linewidth]{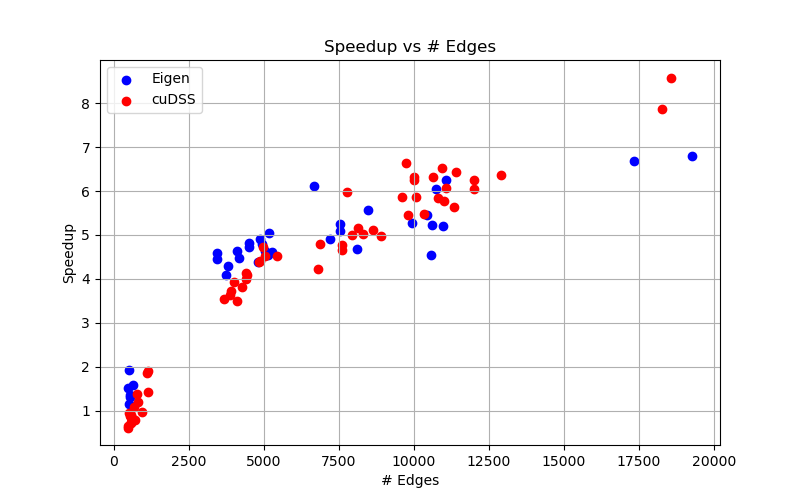}
        \caption{Speedup vs \# of Edges}
        \label{fig:graph_optimization_speedup_b}
    \end{subfigure}
    
    \caption{Speedup of graph optimization versus the number of poses and edges.}
    \label{fig:graph_optimization_speedup}
    \vspace{-15pt}
\end{figure}

\SYS{} exhibits significant performance gains, particularly as the graph size increases. \autoref{tab:combined_results} reports results on sequences from EuRoC and TUM-VI (runtimes include host-to-device and device-to-host transfer and synchronization overhead). 
As reported in \autoref{tab:combined_results}, \SYS{} achieves speedups of up to $3.7\times$ in the overall loop closing pipeline, though improvement varies by sequence: Outdoors7 reaches $3.5\times$, while Room4 shows only $1.2\times$ due to its smaller graph size. Some sequences, such as V103, show no improvement at all, as small graph size lets initial data transfer overhead outweigh the computational gains. \autoref{fig:graph_optimization_speedup} provides further evidence supporting this observation. It shows that the graph optimization speedup grows with the number of poses and edges, and the speedup remains limited for smaller graphs, as discussed further in \autoref{eval:graph_optimization}.

As shown in \autoref{tab:combined_results}, the Absolute Trajectory Error (ATE) RMSE is reported in the last column to compare \SYS{} with ORB-SLAM3. The results show that both systems achieve comparable accuracy, indicating that our proposed optimizations preserve trajectory accuracy and do not degrade the overall performance of the system.
\autoref{fig:room3_trajectory_comparison} illustrates this for the Room3 sequence, showing close alignment between the two estimated trajectories and the ground truth.

\subsection{Individual Component Performances}
\label{eval:individual_performance}


In the rest of this section, we present a detailed analysis of the performance improvements contributed by each optimization incorporated into \SYS{}, highlighting their individual impact on the overall system efficiency.

\subsubsection{Region Detection}
The GPU optimization of the Region Detection component yields an average speedup of $1.1\times$ on the \TUMVI{} dataset in the desktop setting. In the Jetson setting, it achieves an average speedup of $1.1\times$ on both \EUROC{} and \TUMVI{} datasets. Across all sequences, the speedup of this component remains relatively consistent, ranging between $0.9\times$ and $1.2\times$, as shown in \autoref{tab:combined_results}. This behavior is expected, as the component primarily identifies neighbour keyframes using a bag-of-words representation, whose computational cost remains largely invariant across different sequences.

\subsubsection{Loop Fusion}
Executing the Loop Fusion module on the GPU yields an average speedup of $2.3\times$ on the \EUROC{} and $3.3\times$ on the \TUMVI{} dataset in a desktop environment. On the Jetson platform, the corresponding average speedup is $2.8\times$ for \EUROC{} and $5.3\times$ for \TUMVI{}. Notably, certain sequences, such as Magistrale2, achieve higher performance gains of up to $8.4\times$ on the Jetson. This improvement can be attributed to the larger number of map points involved in the fusion process, which enables \SYS{} to more effectively exploit GPU parallelism by leveraging a higher degree of data-level parallelism.

\subsubsection{Graph Optimization}
\label{eval:graph_optimization}
In the graph optimization stage,
\textit{poses} refer to the keyframe poses stored as nodes in the essential graph, and \textit{edges} refer to the constraints connecting them that are used during essential graph optimization. Since each pose is connected only to a small, bounded subset of other poses rather than to every other pose, the number of edges scales roughly linearly with the number of poses rather than quadratically. Implementing the graph optimization stage to the GPU results in an average speedup of $1.5\times$ on the \EUROC{} dataset and $4.0\times$ on the \TUMVI{} dataset in the desktop setting. On the Jetson platform, the same optimization achieves an average speedup of $1.3\times$ for \EUROC{} and $2.6\times$ for \TUMVI{}. Across the sequences evaluated, graph size ranges from approximately $60$ poses and $450$ edges in the smallest sequences (e.g., V103, Room4) to approximately $2300$ poses and $19000$ edges in the largest (e.g., Outdoors7).
\autoref{fig:graph_optimization_speedup} shows that the speedup increases with the number of poses and edges in the graph, highlighting that larger maps benefit more from GPU acceleration. Moreover, as demonstrated in \autoref{fig:graph_optimization_speedup}, Eigen LDLT shows superior performance compared to cuDSS for smaller graph sizes (when the number of poses and edges is close to zero).


\subsubsection{Loop Correction}
This stage combines the Loop Fusion and Graph Optimization components discussed earlier and integrates them with the remaining modules to construct a loop correction. Our optimizations reduce this cost, achieving an average speedup of $1.5\times$ on the \EUROC{} dataset and $3.1\times$ on the \TUMVI{} dataset in the desktop setting. On the Jetson platform, the same optimization attains an average speedup of $1.4\times$ for \EUROC{} and $2.4\times$ for \TUMVI{}. 

\begin{figure}[t]
    \centering
    \includegraphics[width=0.8\linewidth]{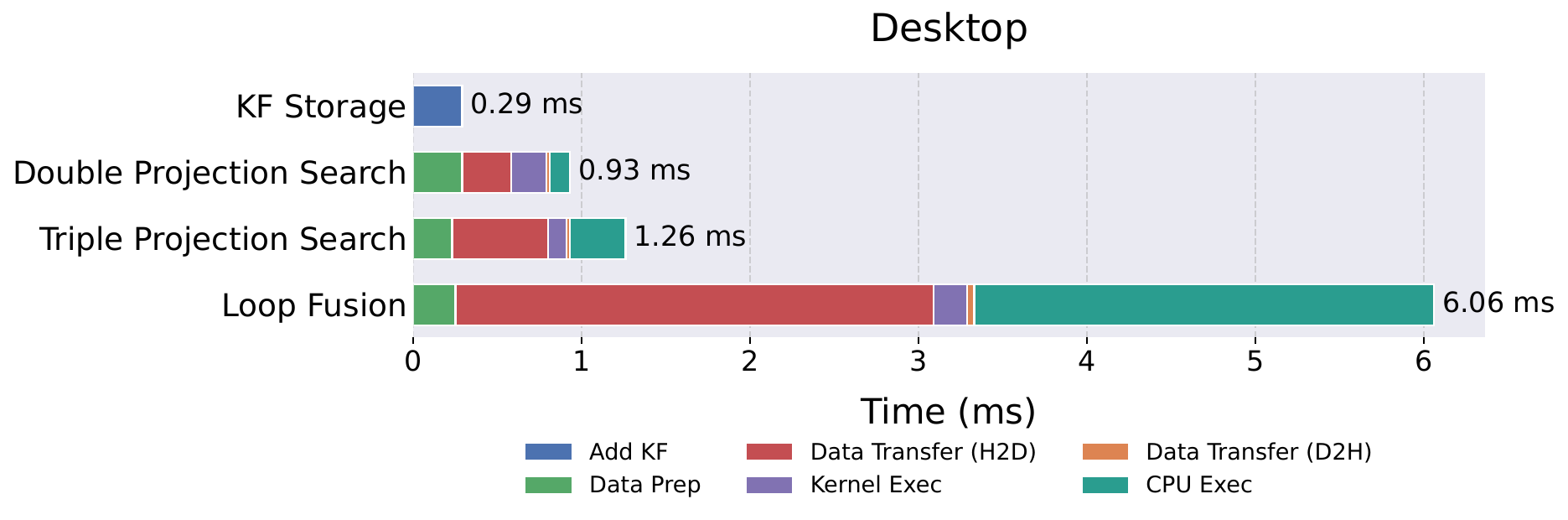}
    \caption{Timing breakdown of Projection Search, Loop Fusion, and keyframe storage in \SYS{} on Corridor1 (desktop).}
    \label{fig:time_analysis}
    \vspace{-10pt}
\end{figure}

\subsection{Projection Search and Loop Fusion Time Analysis}

\autoref{fig:time_analysis} shows the average execution time of the different stages of Projection Search, Loop Fusion, and Keyframe Storage in \SYS{} on the Corridor1, evaluated on the desktop platform. CPU execution time reflects portions retained on the CPU due to negligible latency or unavoidable synchronization and data dependencies. Data preparation time reflects the cost of converting data between CPU- and GPU-compatible formats. Host-to-device transfer dominates execution time for most kernels, driven mainly by the volume of map points and related data. Keyframe transfer is expensive, but occurs only once per keyframe, since keyframes remain resident in GPU memory thereafter.

We also report GPU memory usage to assess feasibility on memory-constrained platforms. \SYS{} preallocates a fixed $\sim$2000\,MiB GPU-resident keyframe buffer (up to 4000 keyframes) once at initialization, plus $\sim$400\,MiB of pinned host memory for CPU--GPU staging. Beyond this, we measured an additional $1000$--$1500$\,MiB of variable overhead across test sequences, from projection search, loop fusion, and data staging. Since keyframe storage is fixed rather than growing per keyframe, \SYS{}'s peak GPU memory usage stays bounded and predictable regardless of sequence length, a desirable property for embedded deployment.


\subsection{Impact on the Rest of the SLAM Pipeline}
\label{eval:impact_on_full_system_performance}


\begin{table}[t]
\vspace{1.0mm}
\centering
\caption{Impact of loop correction.
Local Mapping stop time (ms) and number of keyframes refused by Tracking during LM stop (desktop setting).}
\label{tab:system_impact}
\begin{tabular}{l cc cc}
\toprule
\multirow{2}{*}{Sequence} & \multicolumn{2}{c}{LM Stop Time (ms)} & \multicolumn{2}{c}{KFs Refused} \\
\cmidrule(lr){2-3} \cmidrule(lr){4-5}
 & \SYS{} & Original & \SYS{} & Original \\
\midrule
Magistrale 1 & 74.8  & 252.2 & 2 & 8   \\
Magistrale 2 & 107.1 & 441.0 & 3 & 5  \\
Outdoors 5   & 493.1 & 863.8 & 7 & 15 \\
V203         &   42.9    &   58.1    &  1 &  1  \\
\bottomrule
\end{tabular}
\end{table}

Beyond the loop closing module itself, a detected loop closure also affects the rest of the SLAM pipeline. When a valid loop is found, the Local Mapping pipeline is stopped so that the loop correction can be safely applied to the map, and during this period Tracking refuses to enqueue any new keyframes. Both effects can degrade the quality of the map and the pose estimate immediately following a loop closure, so we evaluate how much \SYS{} reduces this system-wide disruption compared to the original system ORB-SLAM3.

Table~\ref{tab:system_impact} reports, for each sequence, (1) the duration Local Mapping is stopped during loop correction, and (2) the number of keyframes refused by Tracking while Local Mapping is stopped. For example, \SYS{} reduces Local Mapping stop time on Magistrale1 from 252.2\,ms to 74.8\,ms, which directly reduces the number of keyframes rejected by Tracking from 8 to 2.
These results show that \SYS{} reduces disruption to Tracking and Local Mapping.

\begin{table}[t]
\centering
\caption{Tracking latency during Magistrale 1 largest loop closure event, ORB-SLAM3 vs.\ \SYS{}. Normal latency is measured outside loop closure.}
\label{tab:tracking_disruption}
\resizebox{\columnwidth}{!}{%
\begin{tabular}{llcccc}
\toprule
Sequence & System & \makecell{Duration\\(ms)} & \makecell{Frames\\Overlapped} & \makecell{Latency\\During (ms)} & \makecell{Normal\\Latency (ms)} \\
\midrule
\multirow{2}{*}{Magistrale1} & Original & 875 & 16 & 27.0 & 19.3 \\
                             & \SYS{}    & 168 & 3  & 29.0 & 18.3 \\
\midrule
\end{tabular}%
}
\vspace{-15pt}
\end{table}


Beyond Local Mapping stoppage, loop closure also disrupts per-frame tracking latency. For each sequence's largest loop closure event, we compare the latency of overlapping tracking frames to each system's normal (non-overlapping) latency; \autoref{tab:tracking_disruption} reports the results. On Magistrale1, ORB-SLAM3's largest event lasts 875\,ms and overlaps 16 frames; under \SYS{}, the largest event lasts only 168\,ms and overlaps 3 frames. Loop closure slows individual frames under both systems, but by shortening the event itself, \SYS{} substantially cuts the number of frames affected.

\section{Conclusion} \label{sec:conclusion}
In this paper, we introduce \SYS{}, a novel loop closing module for visual SLAM systems that accelerates execution through parallelism strategies and leverages automatic differentiation for graph optimization, thereby addressing major performance bottlenecks in loop closure. \SYS{} exploits both task-level and data-level parallelism on the GPU and adopts a Levenberg–Marquardt pose optimization scheme using the existing GPU-based Graphite framework. In addition, \SYS{} redesigns the data flow to reduce transfer overhead by keeping keyframes entirely in GPU memory. Experiments on both short and long sequences demonstrate that \SYS{} achieves average loop closure speedups of $1.3\times$ on the \EUROC{} dataset and $2.4\times$ on the \TUMVI{} dataset for the embedded platform. On the desktop platform, \SYS{} achieves average speedups of $1.4\times$ and $3.0\times$ on the \EUROC{} and \TUMVI{} datasets, respectively. While \SYS{} is implemented in ORB-SLAM3 using CUDA, the underlying data/task-parallelism partitioning strategy is not specific to either and is expected to generalize to other keyframe-based SLAM systems and GPU platforms.










\bibliographystyle{IEEEtran}
\bibliography{IEEEabrv, references}

\end{document}